\documentclass{ecai} 

\usepackage{latexsym}
\usepackage{amssymb}
\usepackage{amsmath}
\usepackage{amsthm}
\usepackage{booktabs}
\usepackage{enumitem}
\usepackage{graphicx}
\usepackage{color}
\usepackage{multirow}
\usepackage{hyperref}
\usepackage{subcaption}

\def\model{HYSTL}

\newcommand{\BibTeX}{B\kern-.05em{\sc i\kern-.025em b}\kern-.08em\TeX}

\begin{document}


\begin{frontmatter}


\paperid{0821} 


\title{Learning A Universal Crime Predictor with Knowledge-guided Hypernetworks}


\author[A]{\fnms{Fidan}~\snm{Karimova}\thanks{Corresponding Authors. Emails: f.karimova@uq.edu.au, tong.chen@uq.edu.au}}
\author[A]{\fnms{Tong}~\snm{Chen}\footnotemark[*]}
\author[B]{\fnms{Yu}~\snm{Yang}}
\author[A]{\fnms{Shazia}~\snm{Sadiq}}

\address[A]{The University of Queensland, Brisbane, Australia}
\address[B]{The Education University of Hong Kong, Hong Kong SAR, China}


\begin{abstract}
  Predicting crimes in urban environments is crucial for public safety, yet existing prediction methods often struggle to align the knowledge across diverse cities that vary dramatically in data availability of specific crime types. We propose \underline{HY}pernetwork-enhanced \underline{S}patial \underline{T}emporal \underline{L}earning (HYSTL), a framework that can effectively train a unified, stronger crime predictor without assuming identical crime types in different cities' records. In HYSTL, instead of parameterising a dedicated predictor per crime type, a hypernetwork is designed to dynamically generate parameters for the prediction function conditioned on the crime type of interest. To bridge the semantic gap between different crime types, a structured crime knowledge graph is built, where the learned representations of crimes are used as the input to the hypernetwork to facilitate parameter generation. As such, when making predictions for each crime type, the predictor is additionally guided by its intricate association with other relevant crime types. Extensive experiments are performed on two cities with non-overlapping crime types, and the results demonstrate HYSTL outperforms state-of-the-art baselines.

\end{abstract}

\end{frontmatter}

\section{Introduction}

In the era of data-driven governance and predictive policing, accurately predicting where and when crimes are likely to occur can fundamentally transform urban safety policies and law enforcement operations. As a typical spatial-temporal prediction task, crime prediction leverages time series of recorded incidents categorised by type and location to uncover spatial-temporal correlations and forecast future occurrences. This capability not only reassures citizens about their safety but also enables governments to implement data-driven strategies that reduce crime rates and associated societal costs \cite{rumi2018crime}. To capture the complex dynamics inherent in crime data, researchers have proposed advanced models based on deep spatial-temporal networks \cite{huang_deepcrime_2018,huang_mist_2019,rayhan_aist_2023,mandalapu_crime_2023} and graph neural networks (GNNs) \cite{wang_hagen_2021,li_spatial-temporal_2022,chen_spatio-temporal_2023}, facilitating accurate crime predictions. 

Despite promising performance from existing methods, they remain predominantly \textit{\textbf{city-specific}} by design, tailored to the unique characteristics of individual urban environments. This city-specific modeling paradigm significantly limits scalability across diverse geographic locations. Moreover, it overlooks valuable opportunities to extract transferable knowledge from multiple cities' crime records, which is particularly valuable in scenarios where some cities suffer from data sparsity or when zero-shot predictions are needed for newly emerging or under-reported crime types \cite{fdez-diaz_target_2022}. In contrast, a unified approach would facilitate knowledge sharing across cities, leveraging common crime patterns while accommodating regional particularities, ultimately providing a more practical and scalable solution than developing separate models for each location. While cross-city prediction has been extensively investigated in domains such as traffic prediction \cite{chen_semantic-fused_2024,liu_multi-scale_2024} and air quality forecasting \cite{wu_meta-learning-based_2023}, the literature specifically addressing cross-city crime prediction remains notably sparse, highlighting a significant research gap. Unlike domains such as traffic flow, crime records from different regions introduce an added layer of semantic complexity -- terminological inconsistencies, legal categorisation differences, and societal context all vary widely across jurisdictions, bringing unsolved challenges in building such a unified crime predictor across cities. 

The first challenge when developing such a unified crime prediction framework is primarily centered around the semantic misalignment in crime data. Crime data formats exhibit significant variation between cities due to differences in terminology, categorisation systems, and reporting standards. For instance, what one jurisdiction labels as "Larceny" might be classified as "Theft" in another, while certain crime categories may exist in some cities but be entirely absent in others.\footnote{Crime data from specific crime agencies can be explored at the \href{https://cde.ucr.cjis.gov/LATEST/webapp/\#/pages/explorer/crime/crime-trend}{FBI's Crime Data Explorer}.} These inconsistencies lead to mismatches between the input and label spaces across datasets, undermining the ability to jointly train and generalise models across cities. As a result, unified models struggle to learn consistent representations of crime patterns, which is essential for identifying meaningful cross-city correlations and making robust predictions. Previous research exploring transferable models for crime prediction \cite{zhao_exploring_2017,bappee_examining_2021} often operates under the unrealistic assumption that different cities record identical crime types. While techniques such as domain adaptation or transfer learning offer potential workarounds, addressing these semantic misalignment typically requires domain-specific preprocessing or manual mapping, significantly complicating the training pipeline \cite{wu_multi-graph_2022}.

Meanwhile, the second critical challenge lies in the heterogeneous nature of crime dynamics themselves. Most existing crime prediction models adopt a one-size-fits-all approach that assumes shared model parameters  (e.g., graph convolution layers) across all crime types, ignoring the fact that different crimes exhibit fundamentally divergent temporal dynamics and spatial diffusion patterns. On the one hand, when training a universal crime predictor, the heavily shared model parameters across crime types significantly limit predictive accuracy and model expressiveness when handling the diverse nature of criminal activities within and across cities. On the other hand, considering the potential diversity of crime types per city, individually parameterising a model for each crime type in the universal crime predictor can quickly become a hurdle in scalability. 

To bridge this gap, we propose \model, a hypernetwork-enhanced spatial-temporal prediction framework specifically designed for cross-city crime prediction.  Our approach overcomes key limitations in existing methods through two innovative components. First, to address the challenge of crime type (label) mismatch across cities, we construct a domain-specific, multi-relational crime knowledge graph (CrimeKG) that encodes fine-grained relationships between crime types and legal hierarchies, enabling semantic alignment across jurisdictions. Leveraging graph embedding backbones such as metapath2vec++ \cite{dong_metapath2vec_2017}, we derive dense vector representations that capture latent semantic similarities, allowing the model to infer patterns from related categories and maintain robust predictions despite inconsistent or incomplete data. Second, we introduce a hypernetwork \cite{ha_hypernetworks_2016, chen_hate_2024} conditioned on the CrimeKG embeddings, which dynamically generates parameter weights tailored to specific crime types. This design enables context-aware parameter modulation, allowing the prediction model to adapt its internal dynamics to the semantic context of each task, thereby improving generalisation across various cities' diverse urban environments. 

By addressing the critical challenges in cross-city crime prediction, our work entails the following contributions:
\begin{itemize}
\item We design a semantically rich crime knowledge graph (CrimeKG) that systematically encodes inter-crime relationships, enabling alignment across heterogeneous city-level datasets and enhancing model adaptability.
\item We propose a hypernetwork for dynamic parameter generation conditioned on the CrimeKG embeddings, enabling scalable and adaptive crime prediction across diverse cities.
\item Our comprehensive evaluation with Chicago and New York City crime datasets demonstrates the state-of-the-art effectiveness of the framework in cross-city crime prediction.
\end{itemize}


\section{Related Work}

Crime prediction has seen remarkable progress with the integration of advanced machine learning techniques, including \textit{deep learning frameworks} \cite{huang_deepcrime_2018,wang_csan_2020,safat_empirical_2021,dakalbab_artificial_2022,rayhan_aist_2023,mandalapu_crime_2023}, \textit{spatial-temporal models} \cite{zhao_modeling_2017,sun_spatial-temporal_2023,du_systematic_2023}, and \textit{GNNs} \cite{jin_addressing_2020,wang_hagen_2021,xia_spatial-temporal_2021,hou_integrated_2022,li_spatial-temporal_2022,chen_spatio-temporal_2023,roshankar_spatio-temporal_2023,tang_spatio-temporal_2023}. More recently, \textit{Large Language Models (LLMs)} have emerged as a promising paradigm, leveraging spatial-temporal embeddings for urban computing tasks \cite{yuan_unist_2024,li_urbangpt_2024}. Despite these advancements, the field faces significant challenges in developing cross-city generalisable models capable of addressing the heterogeneity of urban environments.

\textbf{Cross-City Generalisation in Spatial-Temporal modeling.} The generalisation of spatial-temporal models across different urban environments presents a significant research challenge, particularly in addressing the disparity between data-rich source cities and data-scarce target cities \cite{liu_multi-scale_2024}. To address this challenge, researchers have extensively investigated transfer learning methods \cite{murakami_spatial_2022,ouyang_domain_2023,liu_frequency_2024} as a means to align urban features and transfer learned patterns across cities. Notable advances in this direction include the work of \cite{ouyang_domain_2023}, which introduces a domain-adversarial graph structure to learn transferable representations, and \cite{liu_frequency_2024}, which proposes a frequency-based pre-training approach to enhance cross-city predictions in few-shot scenarios. Despite demonstrating promising efficacy in knowledge transfer, these approaches face inherent limitations in terms of model interpretability and often require substantial computational resources for effective adaptation.

More recently, advances in this field include a multi-scale traffic pattern bank that captures granular spatial-temporal features \cite{liu_multi-scale_2024} and semantic alignment mechanisms designed to bridge urban semantic disparities \cite{chen_semantic-fused_2024}. While these frameworks offer nuanced perspectives for cross-city prediction, they demand extensive feature engineering and are computationally intensive, limiting their scalability for a diversity of large metropolitan cities. 

While most prior methods in cross-city generalisation have focused on other spatial-temporal tasks like traffic prediction \cite{jiang2024memoryenhancedinvariantpromptlearning, jiang2024physicsguidedactivesamplereweighting}, to the best of our knowledge, few works \cite{zhao_exploring_2017,bappee_examining_2021} have specifically addressed crime prediction, leaving a significant gap in this critical domain. Unlike existing methods that often rely on heuristic alignment or task-specific tuning, our proposed framework, \model, uses a hypernetwork to dynamically adapt spatial-temporal GNN parameters and integrates CrimeKG to enhance knowledge transfer between similar crime types. As such, \model achieves accurate results with cross-city generalisability.


\section{Preliminaries}

\noindent \textbf{Spatial-Temporal Graph.} From the crime data, we construct a spatial-temporal graph denoted as \(G = (V, E, A, X)\), where \(V\) is the set of nodes representing \(R\) geographical regions, with \(|V| = R\); \(E\) is the set of edges capturing spatial relationships between the regions; $A \in \mathbb{R}^{R \times R}$ is the adjacency matrix encoding the spatial connectivity between regions; and \(X \in \mathbb{R}^{T \times R \times C}\) is the feature matrix representing historical crime counts. Each region corresponds to a $3km\times3km$ area within the city, as described in \cite{xia_spatial-temporal_2021}. The crime data spans \(T\) historical days and includes \(C\) crime types. In tensor \( \mathbf{X} \), each entry \( \mathbf{X}_{t, r, c} \) indicates the quantity of criminal incidents of category \( c \) in region \( r \) at time \( t \).

\noindent \textbf{Problem Statement.} The task of crime prediction is to predict future crime counts based on the spatial-temporal graph \(G(V, E, A, X)\). Specifically, the objective is to learn a function \(f(\cdot)\) that takes \(G\) as input and predicts crime counts for the next time step $T+1$. The output is represented as \( \hat{X}_{_{T+1}}\), where $\hat{X}_{T+1} \in \mathbb{R}^{R \times C}$ contains the predicted crime counts for each region and crime type at time \(T+1\). Formally, this is expressed as:
\begin{equation}
f: G(V, E, A, X_{T}) \rightarrow \hat{X}_{T+1},\label{1}
\end{equation}
where the goal is to minimise the discrepancy between the predicted crime counts \(\hat{X}_{T+1}\) and the ground truth crime counts \(X_{T+1}\).


\section{Methodology}

\begin{figure*}[t]
    \centering
    \includegraphics[width=\textwidth, clip]{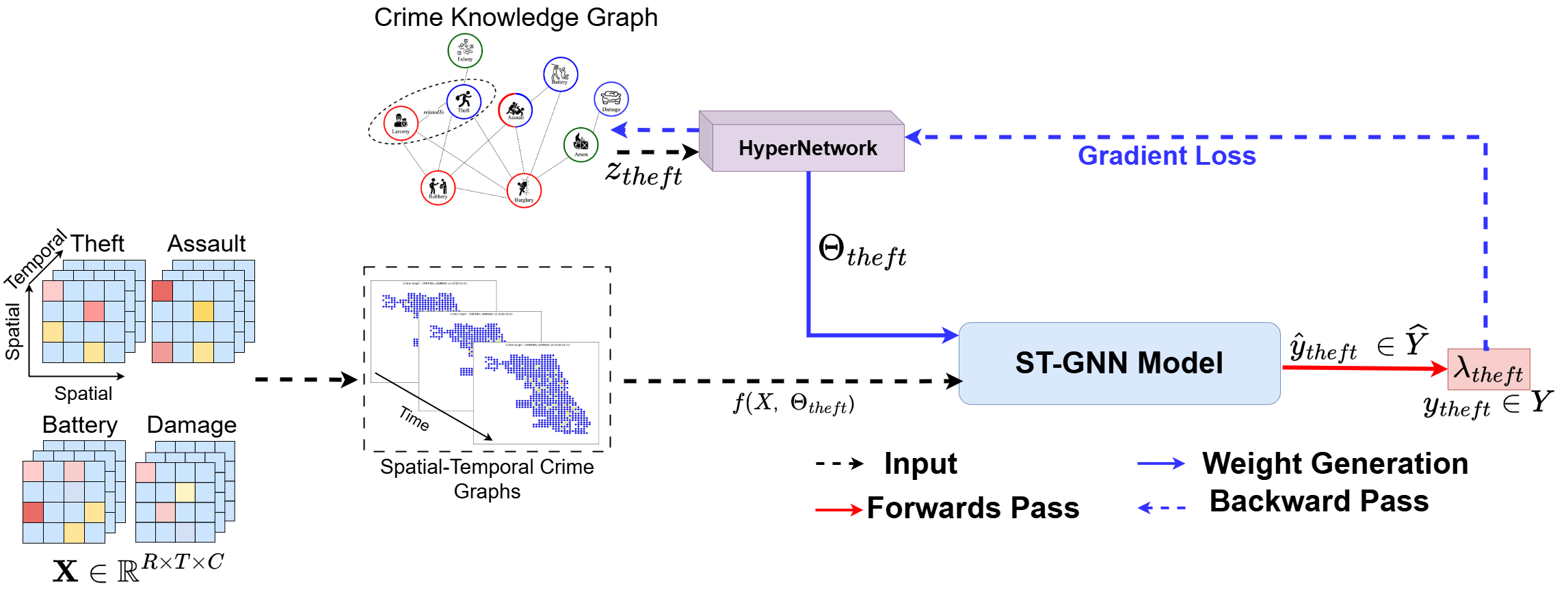}
    \caption{The framework combines static temporal graphs representing crime data with embeddings from CrimeKG to enhance prediction. Each crime type generates specific parameters using a hypernetwork. ST-GNN Model extracts the temporal-spatial features, which are aggregated into context vectors for crime type-specific predictions.}
    \label{fig:framework}
\end{figure*}

We provide a graphical view of \model's workflow in Figure~\ref{fig:framework}, of which the design details are presented in this section.

\subsection{Crime Graph Construction}
We utilise the geographical information of the given city to discretise the spatial domain into a grid structure, such that each grid cell corresponds to a spatial node in the graph. To achieve this, we divide a city of interest into equally spaced grid cells by $3km\times3km$. Each crime record is then mapped to the grid cell, corresponding to its latitude and longitude, converting it into a node \( v_i \) in the spatial-temporal graph. Once the crime data has been mapped to the grid, we proceed to construct the spatial-temporal graph. 

Each graph \( G_{t,c} \) is constructed for a specific time slice and crime type as described above. The graph sequence captures both the spatial-temporal dependencies by incorporating historical data, enabling the model to capture evolving crime patterns over time. This formulation ensures that, at any given time \( t \), the graph \( G_t \) provides a snapshot of the spatial distribution of crime occurrences, while the temporal evolution is captured across multiple time steps. The graph remains structurally consistent across time, allowing for efficient aggregation of spatial and temporal information using GNN methods. To integrate the temporal dimension, we construct a sequence of graphs that evolve over time. Each temporal graph \( G_t = (V, E,A, X_t) \) corresponds to a specific time slice \( t \), where \( X_t \) contains node features for time \( t \). Eventullay, for each crime type \( c \), we can obtain a temporal graph sequence:
\begin{equation}
\mathcal{G}_c = \{G_{t,c} = (V, E, A, \mathbf{X}_{t,c})\}_{t=1}^T. \label{2}
\end{equation}



\subsection{Knowledge Graph Construction}

To address the semantic discrepancy among different cities' crime records and guide the hypernetwork's parameter generation process, we construct a comprehensive crime knowledge graph (CrimeKG) using publicly available resources. To the best of our knowledge, no existing open-source KG provides extensive and generalised relationships among diverse crime types for crime prediction tasks. While frameworks such as \cite{ning_uukg_nodate} offer city-specific KGs, they are limited by the underlying datasets' regional constraints and scope. Inspired by \cite{mernyei_wiki-cs_2022}, we leverage the Wikimedia REST API\footnote{For more details on accessing data via the Wikimedia REST API, see \textit{Wikidata: Data access}, available at \url{https://www.wikidata.org/wiki/Wikidata:Data\_access}.} to extract structured metadata from Wikipedia, ensuring a strong foundation for our KG.

Using the Wikimedia REST API, we retrieved metadata for webpages associated with crime-related concepts. Pages were initially filtered using crime-related keywords (\textit{``crime''}, \textit{``theft''}, \textit{``felony''}, \textit{``criminal law''}, \textit{``offense''}) and then recursively expanded via Wikipedia links. Irrelevant entities (e.g., films or cultural references) were removed to ensure domain relevance, and high-degree nodes were validated to avoid noisy hubs. The extracted metadata included the following key properties: the Wikipedia page title; the pageID, a unique identifier for each Wikipedia page; the language of the page; links to pages related to the current entity, as indicated by Wikipedia's internal relevance ranking; and a concise textual description with an associated image for context.

The goal was to construct a KG encompassing diverse crime-related entities. These entities included crime types, such as fundamental categories like "Theft" and "Robbery"; legal acts and codes, which represent jurisdiction-specific regulations connected to crime types (e.g., "Theft Act 1978" or "California Penal Code"); and associated objects, which refer to items often linked to particular crime categories, such as tools or weapons. In the constructed KG, each node corresponds to an entity (e.g., crime type or legal act), and its metadata properties form the attributes. Edges between nodes are derived based on relationships extracted from Wikipedia's ``relevant pages'' metadata, representing links such as \textit{relatedTo} or \textit{definedBy}. The final CrimeKG includes 3,068 nodes and 5,009 edges, with an average degree of 3.27 and an average eigenvector centrality of 0.0045, providing a semantically informative structure for downstream learning.

Formally, the CrimeKG $\mathcal{G_k}$ represents facts using a set of triplets $\{(h, r, t) \mid h, t \in \mathcal{E}, r \in \mathcal{R}\}$, where $\mathcal{E}$ and $\mathcal{R}$ denote the sets of entities and relations, respectively \cite{zhang_start_2024}. Each triplet $(h, r, t)$ signifies that a relation $r$ exists between the head entity \(h\) and the tail entity \(t\). In the case of CrimeKG, an instance of our triplet would be ("Theft", "\textit{relatedTo}", "Larceny"). 

Given the heterogeneous nature of the constructed CrimeKG, we employ metapath2vec++ \cite{dong_metapath2vec_2017} to learn low-dimensional embeddings for nodes. This method captures both structural and semantic correlations within the CrimeKG by generating random walks (i.e., metapaths) over   entity nodes. The embedding process involves generating a sequence of random walks guided by predefined metapaths. Each node in $\mathcal{E}$ is then represented as a \( d \)-dimensional vector, resulting in an embedding matrix \( \mathbf{Z} \in \mathbb{R}^{|\mathcal{E}| \times d} \). For example, the embedding of a node corresponding to ``Theft'' is denoted as \( \mathbf{z}_{\text{Theft}} \), where \( d \) is the embedding dimension. In particular, we found two metapaths to be most effective: \textit{crime\_type $\rightarrow$ legal\_code $\rightarrow$ crime\_type} and \textit{crime\_type $\rightarrow$ object $\rightarrow$ crime\_type}. Among relation types, \textit{relatedTo} was observed to play the most influential role in connecting diverse entities that share behavioral traits, enabling better generalisation across cities with non-overlapping labels.

The embeddings of specific crime types in the CrimeKG will subsequently serve as inputs to the hypernetwork in the proposed HYSTL framework. It is important to note that the CrimeKG embeddings will be further updated during training as part of the learning process. Through backpropagation, the node embeddings are updated based on prediction errors, enabling the CrimeKG to adapt and refine its representations. 

\subsection{Adaptive Hypernetwork for Target-Specific Parameter Generation}

Given the diversity of crime types, a one-size-fits-all approach to weight generation would fail to capture the nuances specific to each crime type. Instead, we propose using a hypernetwork, which adapts the model's parameters based on the input node embedding corresponding to the target crime type. The adaptive hypernetwork plays a pivotal role in our crime prediction framework, enabling the model to dynamically adjust to each target crime type's specific characteristics. At the core of the adaptive hypernetwork is the ability to generate weights and biases for the predictive model in a target-specific manner. The node embedding $\mathbf{z}_{c} \in \mathbb{R}^d$ of the target crime type $c$ serves as input to the hypernetwork $h(\cdot)$. The embedding $\mathbf{z}_{c}$ encapsulates the rich semantic and contextual information of each crime type from the knowledge graph, enabling the hypernetwork to generate specialised parameters that allow the predictive model to adapt dynamically to the unique patterns and temporal variations of each crime type.

The hypernetwork $h({\cdot})$ is implemented as a simple-yet-effective multi-layer perceptron (MLP), and is designed to generate the parameters for the target-specific predictive GNN model. Specifically, for each target crime type $\mathbf{z_{c}}$, the hypernetwork outputs the corresponding weights $\mathbf{W}_{c}$ and biases $\mathbf{b}_{c}$:
\begin{equation}    
\Theta_c=[\mathbf{W}_{c}, \mathbf{b}_{c}] = h(\mathbf{z_{c}}),  \quad 
\mathbf{W}_{c} \in \mathbb{R}^{d_{\text{in}} \times d_{\text{out}}},\quad 
\mathbf{b}_{c} \in \mathbb{R}^{d_{\text{out}}}, \label{4}
\end{equation}
where $d_{in}$ is the input dimension of the predictive GNN layer (e.g., the dimension of node features), and $d_{out}$ is the output dimension of that layer. For convenience, we denote the generated crime-specific parameters as $\Theta_c$. This setup allows the predictive model to be dynamically instantiated for each crime type using a shared hypernetwork, promoting parameter efficiency and reducing overfitting. Furthermore, it ensures that crime types with similar embeddings (i.e., semantically related crimes in the knowledge graph) will yield similar parameters, thereby encouraging smooth inductive bias over the space of crime types and facilitating efficient knowledge transfer.
For instance, consider two crime types $c_1$ and $c_2$ with related embeddings $\mathbf{z_{c_1}}$ and $\mathbf{z_{c_2}}$. The generated parameters $\Theta_{c_1}$ and $\Theta_{c_2}$ will also be similar, enabling the predictive model for $c_2$ to benefit from learning patterns associated with $c_1$. Compared to standard parameter-sharing approaches like hard parameter tying or multitask learning with shared backbones, the hypernetwork offers a more flexible and continuous form of sharing.

In short, HYSTL offers a fundamental advantage by using hypernetworks in conjunction with knowledge graphs: it transforms CrimeKG embeddings into active controllers of model behavior. As a result, the proposed hypernetwork-based framework is both context-aware (by leveraging the semantic information in the CrimeKG) and structurally scalable (by enabling crime-type-specific parameterisation of the predictive model).

\subsection{GNN-based Crime Predictor}

In this framework, we define the core crime prediction model by extending the graph-based spatial-temporal approach to integrate the outputs from the CrimeKG and hypernetwork. The main predictor is built upon a well-established GNN, namely A3TGCN \cite{zhu_a3t-gcn_2020}. In A3TGCN, nodes within the spatial-temporal graph $G$ are encoded into embeddings, where temporal aggregation and attention-based aggregation are incorporated to capture the dynamic, time-varying dependencies. Then, those intermediate node embeddings are passed through a prediction function to predict the final crime counts. 

Specifically, the A3TGCN model generates the node embeddings $\mathbf{H}_{c,t}$ for each crime type $c$ at time $t$, based on the input feature matrix $\mathbf{X}_{c,t}$ and the adjacency matrix $A$. Notably, the model's parameters $\Theta_{c}$ are generated by the hypernetwork based on the CrimeKG embeddings $\mathbf{z}_c$. This is expressed as:
\begin{equation}
\mathbf{H}_{c,t} = \text{A3TGCN}(\mathbf{X}_{c,t}, A, \Theta_{c}), \label{7}
\end{equation}
where we describe further details below.

\textbf{Attention-based Temporal Aggregation.} To capture the temporal dependencies across multiple time steps, we employ an attention-based aggregation. This technique aggregates the node embeddings from past time steps to create a dynamic representation that incorporates the time-varying aspects of crime occurrences. The temporal aggregation is performed as follows:
\begin{equation}
\mathbf{H}_{c, \text{agg}} = \sum_{t=1}^{T} \alpha_{t} \cdot \mathbf{H}_{c,t}, \quad  \sum_{t=1}^{T} \alpha_t = 1, \label{10}
\end{equation}
where $\mathbf{H}_{c,\text{agg}}$ represents the aggregated node embeddings for crime type $c$ at time step $T$, and $\alpha_{t}$ is the attention weight at time $t$, learned to weigh the importance of different past time steps. This allows the model to selectively focus on the most relevant temporal information. By calculating attention weights $\alpha_t$ for each time step $t$, the temporal aggregation accounts for how much influence each past time step $t$ should have on the current prediction. The attention scores are calculated as follows:
\begin{equation}
\alpha_t = \frac{\exp(\text{score}(\mathbf{H}_{c,t}))}{\sum_{t=1}^{T} \exp(\text{score}(\mathbf{H}_{c,t}))}, \quad \text{score}(\mathbf{H}_{c,t}) = \mathbf{W}_a \cdot \mathbf{H}_{c,t} + b_a, \label{11}
\end{equation}
where $\mathbf{W}_a$ and $b_a$ are parameters learned through the attention mechanism, and $\text{score}(\mathbf{H}_{c,t})$ calculates the relevance of each past node embedding $\mathbf{H}_{c,t}$.

\textbf{Predictor Head.} After performing temporal aggregation with attention weights, the aggregated node embeddings $\mathbf{H}_{c,\text{agg}}$ are then passed through the prediction function $\phi$, which maps the node embeddings to the predicted crime counts for each region $r$ and each crime type $c$ at the future time step $T + 1$. The prediction function is defined as:
\begin{equation}
\hat{\mathbf{Y}}_{c,T+1} = \phi(\mathbf{H}_{c,\text{agg}}), \label{8}
\end{equation}
which converts the aggregated node embeddings into a matrix representing the predicted crime quantities across $R$ regions and for crime type $c$ at time $T +1$.

\subsection{Loss Function and Optimisation}

Given that the model is trained using multi-task learning, the objective is to minimise the total loss $\mathcal{L}_{\text{crime}}$, which is computed using the mean squared error (MSE) between the predicted crime counts $\hat{Y}_{T'+1, r, c}$ and the ground truth values $Y_{T'+1, r, c}$ for all $T'=1,2,...,T-1$:

\begin{equation}
\mathcal{L}_{\text{crime}} = \sum_{\forall T'} \sum_{r=1}^{R} \sum_{c=1}^{C} \left( Y_{T'+1, r, c} - \hat{Y}_{T'+1, r, c} \right)^2, \label{9}
\end{equation}
which allows backpropagation through all model components including the A3TGCN, temporal aggregation layer, the hypernetwork, as well as CrimeKG embeddings, allowing the model to optimise the parameters via gradient descent.

\textbf{Updating The CrimeKG Embeddings.}
To enhance the model's predictive accuracy and adaptability across varying crime contexts, the crime type embeddings learned from the CrimeKG are dynamically updated during training. This mechanism enables the embeddings of crime types to evolve in response to prediction errors, capturing not only static semantic relationships but also dynamic, context-dependent patterns in the data. These refined embeddings are then fed into the hypernetwork, facilitating more accurate and context-aware crime prediction across cities with differing crime type distributions, without compromising model stability or convergence. As a result, our framework acts as a unified, knowledge-informed crime prediction model, capable of leveraging semantic structures to improve performance in diverse crime contexts.


\section{Experiments}

In this section, we evaluate our proposed \model framework with extensive experiments on real-life urban crime datasets and answer the following research questions:

\begin{itemize}
    \item \textbf{RQ1:} How does \model perform compared to various state-of-the-art methods?
    \item \textbf{RQ2:} How do the integration of CrimeKG and the hypernetwork mechanism contribute to the predictive performance of \model?
    \item \textbf{RQ3:} What is the influence of hyperparameters in \model? 
    \item \textbf{RQ4:} How transferable is our \model to different prediction backbones?
\end{itemize}

In the next subsection, we begin by outlining our experimental setup, followed by the description of the evaluation results related to the research questions mentioned above.

\subsection{Experimental Setting}
\subsubsection{Dataset Description} 

In our experiments, we use two crime datasets from New York City and Chicago, each containing various types of crime incidents at different locations.  For New York City, the dataset includes crimes such as Robbery and Larceny, while for Chicago, it includes offences like Damage and Assault. The datasets span two years each. The New York City dataset covers the years 2014 and 2015, while the Chicago dataset corresponds to the years 2016 and 2017. We apply a 3km × 3km spatial grid unit to both cities. The target resolution for the prediction period is set to daily. We construct training and testing data sets with a 7:1 ratio along the time dimension. Hyperparameter optimisation is conducted using the validation set, which consists of data from the final 30 days of the training period. The reported performance of the model is the mean value across all days in the test period for each of the methods compared. These experimental settings closely follow those used in \cite{xia_spatial-temporal_2021}, \cite{li_spatial-temporal_2022}. The summary of the data can be found in Table~\ref{tab:crime-cases}. 
\begin{table}
    \centering
    \caption{Crime Cases for New York City and Chicago Datasets}
    \begin{tabular}{lrr}
        \toprule
        City& Crime Type& Number of Offences\\
        \midrule
        New York City& Burglary& 31799\\
                  & Robbery& 33,453\\
 & Assault&40429                      \\
 & Larceny&85899                      \\
 \bottomrule
        Chicago& Theft& 124,630\\
        & Battery& 99,389\\
                  & Damage& 59,886                     \\
 & Assault&37,972                     \\
        \bottomrule
    \end{tabular}
    \label{tab:crime-cases}
\end{table}

\subsubsection{Evaluation Metrics}  

To assess the accuracy of urban crime predicting, we employ two commonly used metrics: Mean Absolute Error (MAE) and Mean Absolute Percentage Error (MAPE). To ensure a fair comparison and mitigate evaluation biases, the prediction performance of all competing methods is averaged over all days in the test period. It is important to note that lower MAE and MAPE values signify better predictive accuracy, underscoring their importance in evaluating model performance.

\subsubsection{Baselines for Comparison}
We evaluate our model against several state-of-the-art methods grouped by methodology. Recurrent Neural Network (RNN)-based models include Deepcrime \cite{huang_deepcrime_2018},  leveraging advanced deep learning models to analyze spatiotemporal patterns and D-LSTM \cite{yu_spatio-temporal_2018} , which enhances temporal modeling with hypernetworks but lacks explicit spatial modeling. Graph-based models such as GWN \cite{wu_graph_2019}, GMAN \cite{zheng_gman_2020} capture spatial-temporal dependencies using GNNs, with GMAN incorporating multi-head attention to improve spatial-temporal correlations and DMSTGCN \cite{han_dynamic_2021}. Hybrid models include MTGNN \cite{wu_connecting_2020}, which combines GNNs with temporal models for multivariate time-series predicting, and STSHN \cite{xia_spatial-temporal_2021}, which uses hierarchical attention to model short- and long-term dependencies. STHSL \cite{li_spatial-temporal_2022} leverages self-supervised learning for robust spatial-temporal representation, while CL4ST \cite{tang_spatio-temporal_2023} applies contrastive learning for improved feature extraction in noisy data, and HCL \cite{liang_hawkes-enhanced_2024}  combines hypernetworks with contrastive learning for adaptable and robust predictions. Further details are provided in the related works section.

Additionally, we compare our model against UrbanGPT \cite{li_urbangpt_2024}, a recent framework designed for cross-city urban prediction. Unlike most methods that assume consistent crime types across cities, UrbanGPT emphasises knowledge transferability, making it a particularly relevant benchmark for our setting. This comparison is crucial, as it evaluates how well HYSTL handles differing crime type distributions across cities.

\subsubsection{Implementation Details}

The spatial-temporal crime prediction model utilises several hyperparameters to optimise performance and computational efficiency.  The hypernetwork processes embeddings of dimension 16 to generate 44 parameters, comprising 32 weights for the Temporal GNN and 12 biases for the linear layer. The Temporal GNN operates with 2 node features as input, produces 32 output channels through its A3TGCN layer. Training is conducted using the Adam optimizer with a learning rate of 0.01 over 20 epochs, processing up to 100 snapshots per epoch to manage the computational load. We implement all codes using PyTorch and PyTorch Geometric library. All experiments are conducted on NVIDIA GeForce RTX 4090 GPUs and Intel Core i7-13700K.  The model and hypernetwork parameters are trained jointly on a CPU, though the framework supports GPU acceleration.  These hyperparameters were chosen based on empirical testing to balance model complexity and prediction accuracy.

\subsection{Performance Comparison (RQ1)}

Our experimental evaluation in Table \ref{tab:mae_mape_results} demonstrates that the proposed model outperforms eight baseline models in predicting crime across multiple categories in New York City and Chicago. It achieves consistently lower MAE scores across all crime types, with improvements ranging from 16.2\% to 57.1\% over recent benchmarks, highlighting a significant advancement in urban crime prediction.

In New York City, our model excels in burglary prediction (MAE = 0.4387), improving by 16.8\% over HCL and by 60.1\% over LSTM. Larceny prediction follows a similar trend, with our model (MAE = 0.4652) outperforming CL4ST (MAE = 0.5651). Notably, the model achieves the lowest MAE for robbery (0.4342) and assault (0.4042) predictions, marking substantial improvements over earlier graph-based models like GWN and MTGNN.

In Chicago, our model maintains strong performance across all crime types. Theft prediction (MAE = 0.5399) shows a large improvement over STHSL (MAE = 1.4139), demonstrating our model's ability to capture complex spatial-temporal dynamics. Battery and assault predictions (MAE = 0.4982, 0.5226) outperform the recent HCL model by 53.0\% and 16.8\%, respectively.

Notably, HYSTL outperforms UrbanGPT, achieving lower MAE and MAPE across all evaluated crime types, which highlights the effectiveness of dynamic knowledge graph updates and shared crime-type embeddings in enabling robust cross-city generalization despite non-overlapping crime categories.

To assess robustness, we perform experiments over multiple runs and report mean values with standard deviations. In New York City, the proposed model achieves $\text{MAE} = 0.436 \pm 0.025$ and $\text{MAPE} = 0.320 \pm 0.038$. In Chicago, the results are $\text{MAE} = 0.518 \pm 0.017$ and $\text{MAPE} = 0.355 \pm 0.041$. The small standard deviations indicate stable convergence and consistent performance across repeated trials.

While our model demonstrates superior MAE performance, it also achieves better generalisation across cities. The consistent error rates across New York City and Chicago suggest robust handling of urban context variations. This is in contrast to baseline models, which exhibit more significant performance gaps between cities.

\begin{table*}[t]
    \centering
    \caption{Performance comparison of different models on crime prediction task across New York City and Chicago}
\resizebox{\textwidth}{!}{
\begin{tabular}{lrlllllllrlllllll}
        \toprule
        Models& \multicolumn{8}{c}{New York City}& \multicolumn{8}{c}{Chicago}\\
 & \multicolumn{2}{c}{Burglary}& \multicolumn{2}{c}{Larceny}& \multicolumn{2}{c}{Robbery}& \multicolumn{2}{c}{Assault}& \multicolumn{2}{c}{Theft}& \multicolumn{2}{c}{Battery}& \multicolumn{2}{c}{Assault}& \multicolumn{2}{c}{Damage}\\
 & MAE& MAPE& MAE& MAPE& MAE& MAPE& MAE& MAPE& MAE& MAPE& MAE& MAPE& MAE& MAPE& MAE&MAPE\\
        \midrule
        LSTM&  1.0993&  1.7265&  0.9814&  1.2176&  0.7468&  1.7527&  0.7365& 1.016&  1.1098&  2.6426&  0.7931&  1.5299&  0.8458&  1.8547&  0.7055& 1.010\\
                  DeepCrime& 0.8227& 0.5508& 1.0618& 0.5351& 0.8841& 0.5537& 0.9222&0.5677& 1.3391& 0.5430& 1.1290 & 0.5389& 0.7737& 0.4616 & 0.9096&0.4960\\
        GWN& 0.8012& 0.5352& 1.0510& 0.5505& 0.8698& 0.5374& 0.8911&0.5684& 1.3198& 0.5609& 1.1401& 0.5611& 0.7504& 0.4582& 0.8622&0.4849\\
        MTGNN& 0.8401& 0.5425& 1.0398& 0.5430& 0.8757& 0.5461& 0.9095&0.5725& 1.3102& 0.5378& 1.1308& 0.5601& 0.7573& 0.4569& 0.8665&0.4855\\
 GMAN& 0.8649& 0.5723 & 1.0602& 0.5440& 0.9333& 0.5716& 0.9341 & 0.5801 & 1.3301& 0.5298& 1.1341& 0.5498& 0.7751& 0.4701& 0.823&0.4939\\
 DMSTGCN& 0.8376& 0.5485& 1.0410& 0.5464& 0.8597& 0.5403& 0.9036 & 0.5601& 1.3292 & 0.5291 & 1.1297 & 0.5552 & 0.8058 & 0.4759& 0.8698 &0.4877\\
 STSHN& 0.5548& 0.3530& 0.5947& 0.3602& 0.6345& 0.3887& 0.5866& 0.3836& 0.6747& 0.4116& 0.6728& 0.4332& 0.6483& 0.5000& 0.7347&0.5058\\
 STHSL& 0.7268& 0.4716& 1.0529& 0.5136& 0.7980& 0.4621& 0.8463& 0.4930& 1.4139& 0.4832& 1.1235& 0.4954& 0.7917& 0.4698& 0.8900&0.4580\\
                  CL4ST& 0.5272& 0.3468& 0.5651& \textbf{0.3539}& 0.6030& 0.3819& 0.5573&0.3769& 0.6571& 0.4008& 0.6553& 0.4218& 0.6314& 0.4100& 0.7156&0.5039\\
                   HCL& 0.5228& 0.3146& 0.9017& 0.4142& 0.6116 & 0.3391 & 0.6487 & 0.3662& 1.2506& 0.4501& 1.0601& 0.4621& 0.6284& 0.3712& 0.8006&0.4269\\
                UrbanGPT& 0.4712& 0.6531& 0.5121& 0.6645& 0.4913& 0.6113& 0.4811& 0.5321& 0.5521& 0.6364& 0.5125& 0.5675& 0.5943& 0.6123& 0.5921& 0.6567\\
                MVST&  0.5534& 0.3123& 0.7544& 0.4211& 0.6521& 0.3454& 0.6732& 0.3154& 0.7689& 0.3578& 0.8451& 0.5517& 0.6987& 0.4345& 0.7521&0.4235\\
 \bottomrule
 Ours& \textbf{0.4387} & \textbf{0.2810} & \textbf{0.4652} & 0.3699 & \textbf{0.4342} & \textbf{0.3029} & \textbf{0.4042} & \textbf{0.3233} & \textbf{0.5399} & \textbf{0.3315} & \textbf{0.4982} & \textbf{0.3813} & \textbf{0.5226} & \textbf{0.3064} & \textbf{0.5145} &\textbf{0.4020} \\
        \bottomrule
    \end{tabular}}
    \vspace{0.2cm}
    \label{tab:mae_mape_results}
    \vspace{0.5cm}
\end{table*}

\subsection{Model Ablation Study (RQ2)}

\begin{figure}
    \centering
    \begin{subfigure}{\columnwidth}
        \centering
        \includegraphics[width=\columnwidth]{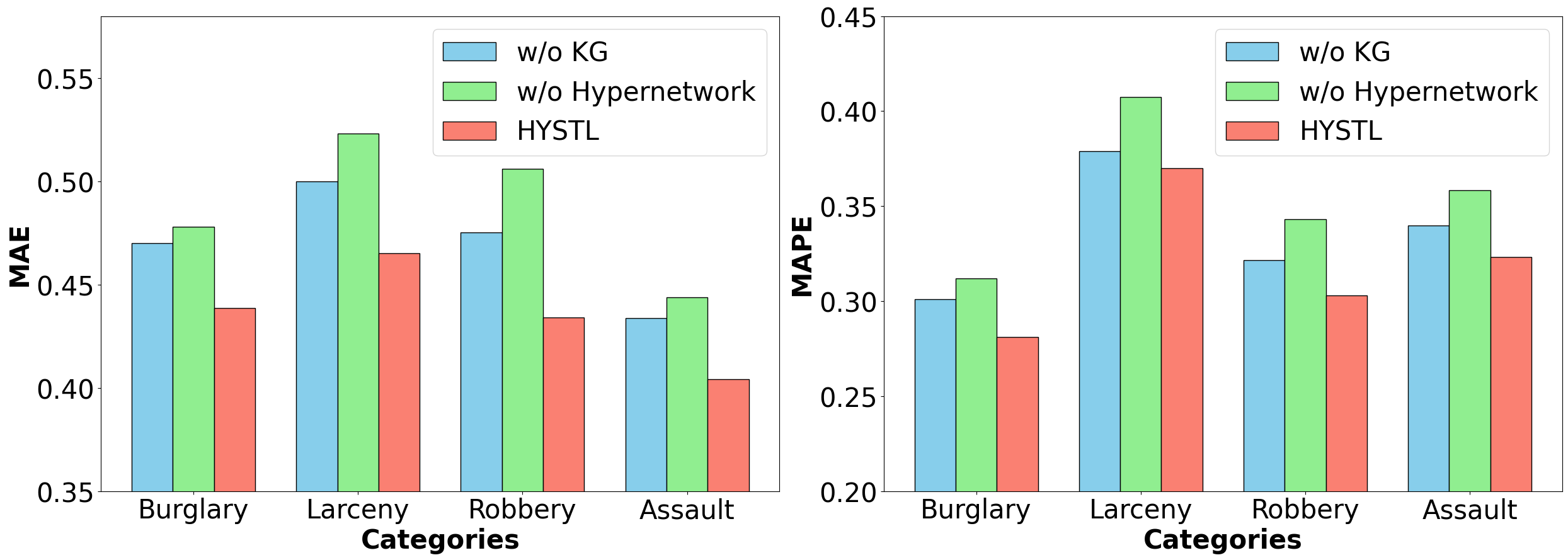}
        \caption{Evaluation Results on New York City Data}
    \vspace{0.5cm}
    \end{subfigure}
    \begin{subfigure}{\columnwidth}
        \centering
        \includegraphics[width=\columnwidth]{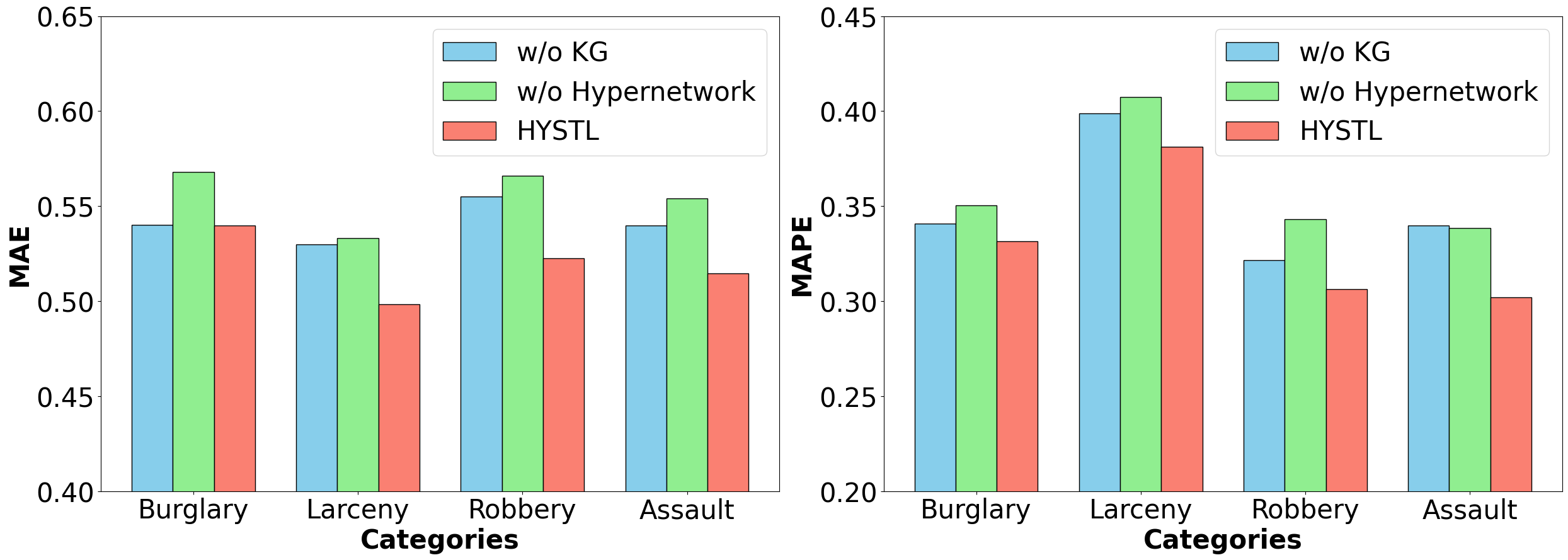}
        \caption{Evaluation Results on Chicago Data}
    \end{subfigure}
    
    \vspace{0.5cm}
    \caption{Ablation study of \model\ framework, in terms of \textit{MAE} and \textit{MAPE}.}
    \label{fig:ablation}
    \vspace{0.5cm}
\end{figure}

The ablation study in Figure \ref{fig:ablation} demonstrates the effectiveness of the HYSTL framework across four crime types: Burglary, Larceny, Robbery, and Assault, comparing it to two baseline variants, one without knowledge graphs (w/o CrimeKG) and one without hypernetwork components (w/o Hypernetwork).

In the "w/o CrimeKG" variant, entity embeddings are initialised using GloVe \cite{pennington_glove_2014}, capturing basic semantic relationships without the structural insights of knowledge graphs. In the "w/o Hypernetwork" variant, knowledge graph embeddings are directly integrated into the GNN, bypassing the hypernetwork's dynamic adaptability.

HYSTL consistently achieves lower MAE scores across both cities, with a significant improvement (15–20\%) in violent crimes (Robbery, Assault) and moderate reductions (5–10\%) in property crimes (Burglary, Larceny) compared to both baselines. These results highlight the importance of knowledge graphs and hypernetwork components in enhancing predictive accuracy, especially for violent crime predictions.

\subsection{Hyperparameter Study (RQ3)}

\begin{figure}[t]
    \includegraphics[width=\columnwidth]{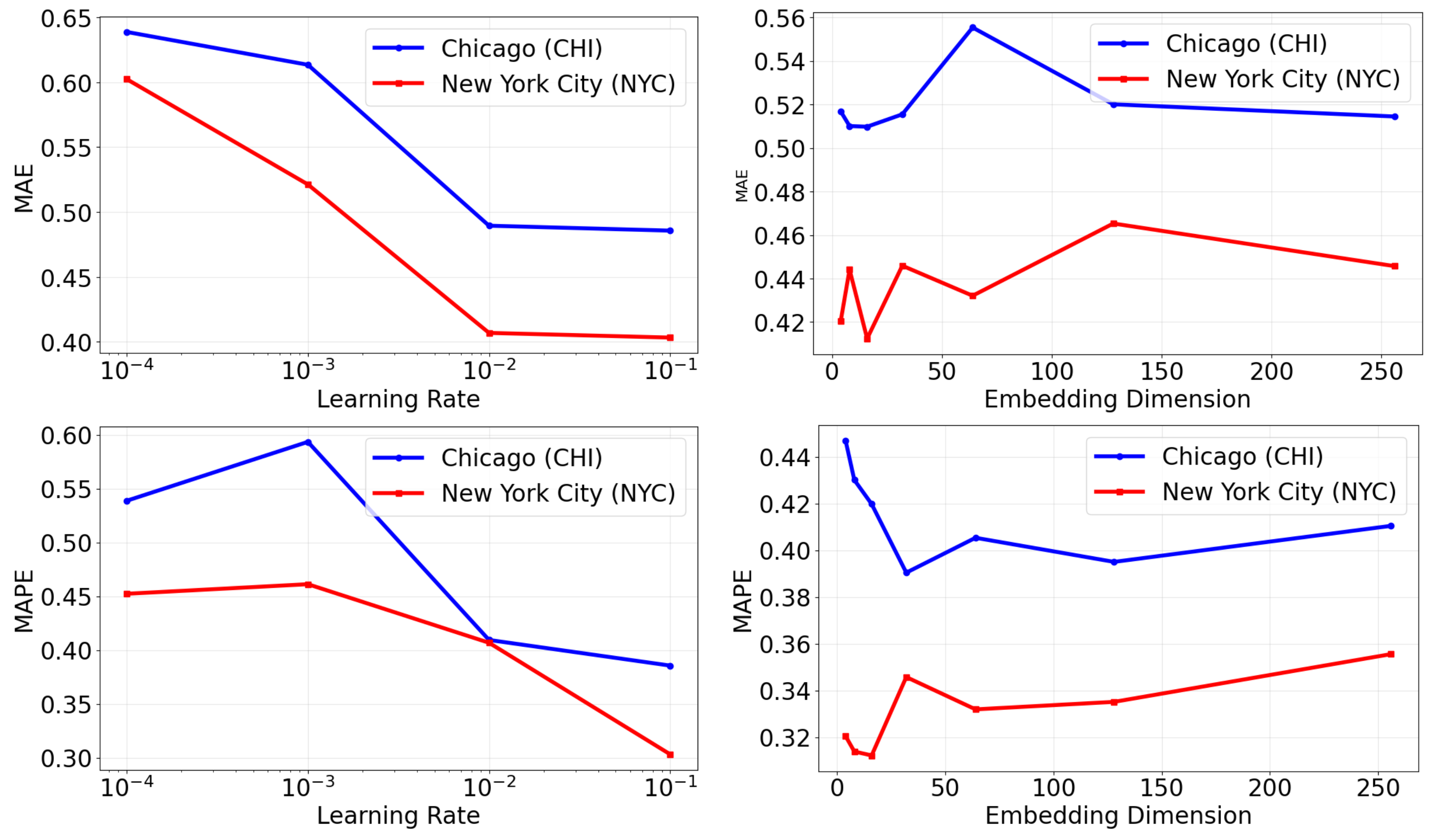}
    \vspace{0.1cm}
    \caption{Impact study for various hyperparameters in \model’s performance on Chicago and New York crime data, in terms of MAE and MAPE.}
    \vspace{0.8cm}  
    \label{fig:hyperparameters}
\end{figure}

We evaluate the impact of key hyperparameters on \model, focusing on embedding dimension and learning rate (Figure \ref{fig:hyperparameters}). An embedding size of 16 achieves the best performance, while varying the embedding size from 4 to 256, accuracy decreases slightly as the dimension increases beyond 16, suggesting that larger embeddings may lead to overfitting, particularly with sparse crime data. A fixed learning rate of 0.01 was empirically selected as optimal, since learning rate scheduling provided only marginal gains and did not justify the added complexity. These results highlight the need for balanced hyperparameter selection to ensure both efficiency and accuracy.



\subsection{Transferability Study (RQ4)}
We evaluate the compatibility of our proposed HYSTL framework with various state-of-the-art GNN architectures, such as T-GCN \cite{zhao_t-gcn_2020}, EGCN-H \cite{li_diffusion_2017}, DCRNN \cite{pareja_evolvegcn_2020} and A3TGCN.

\begin{table}[h!]
    \centering
    \caption{Performance of \model with various backbones}
    \begin{tabular}{c|cc|cc}
        \hline
        \multirow{2}{*}{Models} & \multicolumn{2}{c|}{New York City} & \multicolumn{2}{c}{Chicago} \\ \cline{2-5} 
                                & MAE& MAPE& MAE& MAPE\\ \hline
        A3TGCN & 0.4356& 0.3193& 0.5188& 0.3553\\
 T-GCN & 0.4431& 0.3205& 0.5197&0.3611\\
 EGCN-H & 0.4601& 0.3510& 0.5446&0.3863\\ 
        DCRNN & 0.4472& 0.3313& 0.5355& 0.3727\\
        \hline
    \end{tabular}
    \label{tab:transferability}
\end{table}
 We use A3TGCN as the baseline model for all experiments conducted in this study. The experimental results demonstrate that the integration of HYSTL significantly enhances the performance of the models, confirming that our framework is adaptable and effective across different GNN backbones. The results, shown in Table~\ref{tab:transferability}, reveal substantial improvements in predictive accuracy.


\section{Conclusion}

In this study, we propose a novel crime prediction framework that integrates hypernetworks and CrimeKG embeddings to enhance spatial temporal predictions. By dynamically generating model parameters through hypernetworks, our approach effectively captures the complex dependencies in crime data, outperforming existing baselines. This model adapts to varying crime patterns across cities. Our findings demonstrate the potential of combining hypernetworks with graph-based and recurrent architectures to tackle challenges such as data sparsity, temporal fluctuations, and spatial relationships between crime hotspots. Future work could extend this approach to real-time crime prediction systems and other domains, such as disaster management and urban planning. As a future direction, we aim to extend this approach to zero-shot \cite{qin_generative_2020} and few-shot \cite{liu_multi-scale_2024} scenarios, enabling it to handle cases with minimal labelled data effectively, as well as, explore the interpretability \cite{tang_explainable_2023} of our framework, providing deeper insights into its predictions and making it more accessible for decision-makers in law enforcement and public safety.



\begin{ack}

F. Karimova, T. Chen, and S. Sadiq are supported by the Australian Research Council under the streams of Discovery Project (No. DP240101814), Discovery Early Career Research Award (No. DE230101033), Linkage Project (No. LP230200892 and LP240200546), and Industrial Transformation Training Centre (No. IC200100022). Y. Yang is supported by the EdUHK Project under Grant No. RG 67/2024-2025R.

\end{ack}



\bibliography{mybibfile}

\end{document}